\documentclass[runningheads]{llncs}

\usepackage{eccv}

\usepackage{eccvabbrv}

\usepackage{graphicx}
\usepackage{booktabs}

\usepackage[accsupp]{axessibility}  

\usepackage[breaklinks,colorlinks,citecolor=eccvblue]{hyperref}

\usepackage{orcidlink}

\usepackage{enumitem}
\usepackage{amsmath,amsfonts,bm,bbm}
\usepackage{multirow}
\usepackage[table]{xcolor}
\newcommand{\gcell}{\cellcolor{gray!20}\color{darkgray!70}}
\newcommand{\gcellonly}{\cellcolor{gray!20}}
\definecolor{dark_green}{rgb}{0, 0.5, 0}
\newcommand{\ours}{UniD\xspace}

\def\Secref#1{Section~\ref{#1}}

\def\eqref#1{equation~\ref{#1}}

\def\R{\mathbb{R}}

\DeclareMathOperator*{\argmin}{arg\,min}
\newcommand\pad{\vspace{-0.33cm}\\} 
\usepackage{amssymb}
\usepackage{pifont}
\newcommand{\cmark}{\ding{51}}%
\newcommand{\xmark}{\ding{55}}%

\newcommand\projectpage{https://unid-video.github.io}

\newcommand\titlename{Unified Video Dense Prediction from Disjoint Data}

\begin{document}

\title{\texorpdfstring{\makebox[\textwidth]{\titlename}}{\titlename}} 

\titlerunning{\titlename}

\author{
Yihong Sun\inst{1,2,\normalfont*} \and 
Seoung Wug Oh\inst{1} \and
Jiahui Huang\inst{1} \and \\
Bharath Hariharan\inst{2} \and
Joon-Young Lee\inst{1}
}

\authorrunning{Sun et al.}

\institute{$^{1}$Adobe Research \qquad $^{2}$Cornell University}

\maketitle

\renewcommand{\thefootnote}{\fnsymbol{footnote}}
\footnotetext[1]{Work done during an internship at Adobe Research.}
\renewcommand{\thefootnote}{\arabic{footnote}}

\begin{abstract}
Scene understanding requires simultaneous prediction about geometry, appearance, and semantics.
However, existing task-specific annotations are fragmented across incompatible, domain-specific datasets.
Current unified systems circumvent this by restricting training to fully co-annotated data, or by incurring the large computational cost of pseudo-labeling.
To mitigate this, we introduce \textbf{\ours}, a unified video model that jointly predicts eight dense scene properties—depth, surface normals, semantic segmentation, boundaries, human parts, albedo, shading, and materials—all learned from disjoint, domain-specific datasets. 
We propose a simple yet effective distillation step in which per-task experts supervise a unified backbone through lightweight task projectors, eliminating the need for annotation overlap or pseudo-labeling. 
Our key insight is that the strong visual priors of a pretrained diffusion model are sufficient to bridge the domain gaps introduced by disjoint training sources, enabling robust generalization to scene-task combinations never seen during training.
\ours achieves competitive performance against per-task specialists and multi-task baselines, with strong generalization to out-of-distribution scenarios and enhanced temporal and cross-task consistency.
Code and video results are available at \url{\projectpage}.

\end{abstract}

\section{Introduction}
\label{sec:intro}

To understand a scene, a vision system must answer many questions at once: how far away are the visible objects? What materials are they made of? And how does light interact with their surfaces? 
These questions correspond to distinct dense prediction tasks—depth, material segmentation, intrinsic decomposition—each of which has spawned its own specialized model, learning framework, and datasets. 
However, the real world is a combination of all of these properties and ever-changing: an embodied agent navigating or manipulating a scene would need all of these cues simultaneously and continuously.

\begin{figure}[t!]
  \centering
  \includegraphics[width=\textwidth]{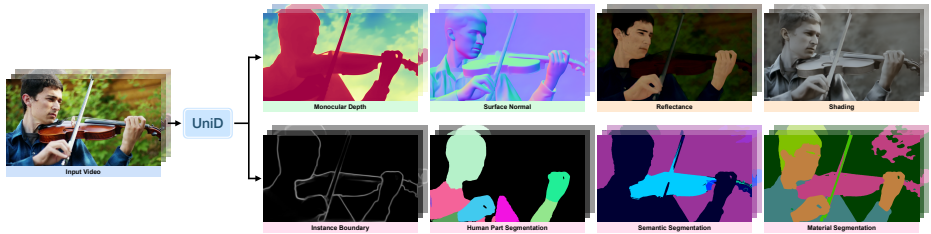}
  \caption{Unified Video Dense Predictions by \ours.}
  \label{fig:teaser}
\end{figure}

This has motivated recent development of \emph{unified} dense prediction models~\cite{invpt,taskprompter,taskexpert,mtmamba,stablemtl,taskdiffusion,diception} that output multi-task predictions jointly.
However, almost all existing methods assume that all tasks are co-annotated on the same images—an assumption that is unrealistic with how task-specific data is collected in practice.
Geometric annotations such as depth and normals can be acquired at scale via stereo or RGB-D cameras, but are typically captured in environments with limited semantic diversity, \eg architectural landscapes.
Semantic annotations require manual labeling over a diverse set of scenes, yielding rich category diversity but rarely paired with geometric ground truth.
Finally, intrinsic decompositions~\cite{barron2014shape, narihira2015direct} are prohibitively expensive to measure in the real world and largely confined to synthetic domains.
Driven by these fundamental differences, forcing these fragmented datasets into a single co-annotated training set is either infeasible or requires large-scale pseudo-labeling that is unscalable: each new task requires re-annotating the entire dataset.

On top of this, extending into the video domain would require a substantial amount of annotated videos. 
While feasible for geometry (\eg RGB-D video captures), this becomes prohibitively expensive for semantics and intrinsics tasks.
As a result, existing video-capable methods are either confined to the self-driving domain~\cite{vtdnet} or focus mainly on low-level geometric estimations~\cite{l4p}.

These constraints raise a natural question: \emph{can we learn unified dense prediction for both images and videos from disjoint, domain-specific datasets?}
We present \textbf{\ours}, a \textbf{uni}fied video model learned from \textbf{d}isjoint, domain-specific data that jointly estimates eight dense scene properties—depth, surface normals, semantic segmentation, boundaries, human parts, albedo, shading, and materials (Figure~\ref{fig:teaser}).
In order to bridge the domain gaps arising from disjoint training sources, we first learn a task-specific latent space with strong generalization to unseen examples.
We achieve this generalization by leveraging the visual priors of a diffusion model that is pretrained on internet-scale data~\cite{rombach2022high}.
Then, we train a unified model to reconstruct all task-specific latent embeddings for any input, via a lightweight per-task latent projector that maps a shared backbone representation into each task's latent space.
%
%
By computing the task-specific embeddings on-the-fly, the unified training requires no co-annotated data or pseudo-labels.

\ours offers several key advantages. 
First, it achieves enhanced generalization and temporal stability.
By jointly leveraging disparate sources across diverse distributions---such as synthetic and real-world, indoor and outdoor, and static images alongside dynamic videos---the model avoids overfitting and demonstrates significant robustness to out-of-distribution (OOD) scenarios. 
Crucially, tasks without video supervision inherit temporal coherence from other video-centric tasks, ensuring stable and consistent predictions over time.

Second, the unified representation provided by a shared backbone ensures both cross-task consistency and high inference efficiency. 
By predicting a single unified feature manifold, our model enforces strong structural alignment across tasks.
This leads to spatially congruent predictions across different tasks---such as consistent boundary delineations---which are often misaligned in independent task-specific models. 
Furthermore, this shared architecture significantly reduces computational costs, as the backbone features are computed only once to support multiple task-specific projectors, incurring minimal per-task overhead.

To demonstrate these advantages, we conducted extensive experiments across all tasks, where \ours achieves competitive performance compared to the existing task-specific models.
Furthermore, \ours demonstrates superior OOD generalization, temporal stability, and cross-task consistency over existing unified models and a multi-task baseline utilizing a frozen DINOv3-H backbone, proving its effectiveness in complex, real-world video dense prediction.



\section{Related Work}
\label{sec:related}

\paragraph{Multi-task dense prediction.}
Unifying dense prediction tasks within a single model improves efficiency and encourages cross-task complementarity~\cite{taskonomy}.
Naturally, many existing works~\cite{invpt,invpt++,mtmamba,mtmamba++,taskprompter,pgt,mqtransformer,demt,taskexpert,mlore} acquire multi-task capabilities by requiring every training image to be annotated for every task.
To relax this constraint, recent works~\cite{li2022learning,diffusionmtl} also investigated learning with partial per-image annotations to improve label efficiency.
However, it still assumes that each image contains a subset of labeled tasks, drawn from the same co-annotation distribution as before.
In addition, few-shot generalists~\cite{vtm,chameleon,xia2025ideal} reduce per-task annotation requirements through meta-learning and episodic training.
However, they still rely on shared meta-datasets across tasks and do not take full advantage of large-scale task-specific datasets.

In parallel, several works adopt pseudo-labeling to enforce large-scale annotation co-occurrence~\cite{multimae,vtdnet,udpdiff,bachmann20244m,diception}.
While effective, pseudo-labeling is costly, and the entire labeling pipeline must be re-run when a new task or dataset is added.
Notably, 4M-21~\cite{bachmann20244m} further requires discrete tokenization of all task outputs, and its masked modeling objective couples all tasks into a 
uniform token prediction problem, precluding task-specific optimization objectives.
In contrast, \ours trains on disjoint, task-specific datasets while allowing flexible task-specific learning objectives, with no annotation co-occurrence required.

\paragraph{Diffusion models for dense prediction.}
Diffusion models~\cite{ho2020denoising,song2020score,rombach2022high,esser2024scaling} have shown impressive image generation capabilities.
Their rich visual priors learned from large-scale pretraining have enabled many applications, including conditional generation~\cite{ho2022classifier,ye2023ip,zhang2023adding,mou2024t2i}, editing~\cite{nichol2021glide,kawar2023imagic,brooks2023instructpix2pix}, and video generation~\cite{ho2022video,ho2022imagen,svd,videopoet}.

Following this, recent methods~\cite{xu2024matters,he2024lotus,ye2024stablenormal,marigold,zhu2024unleashing,tang2023emergent} propose to leverage the generative priors to learn dense predictions (\eg depth and surface normals) that generalize beyond the limited training data.
In addition to these task-specialists, direct multi-task training with diffusion models have been proposed with either fully-annotated data~\cite{taskdiffusion}, pseudo-labeling~\cite{diception}, or synthetic-only supervision~\cite{stablemtl}.
Notably, DICEPTION~\cite{diception} unifies multi-tasks by mapping all into RGB space with conditioned task prompts (\eg ``image2depth'').
However, the RGB unification cannot naively represent tasks with high-dimensional outputs (\eg semantic segmentation), and text-conditioned generation requires $K$ separate forward passes to obtain all $K$ predictions, precluding simultaneous multi-task inference.

\ours instead leverages diffusion models to learn a task-specific latent space, enabling per-pixel, task-specific optimization objectives, while leveraging the generative priors of the diffusion backbone to bridge domain gaps from disjoint training sources.
Architecturally, rather than iterative denoising, \ours adopts a single-step diffusion with a shared backbone and lightweight per-task latent projectors, producing all task predictions simultaneously in a single forward pass.

\paragraph{Video dense prediction.}
Dense prediction has been studied extensively in the video domain, including video depth estimation~\cite{chronodepth,yang2024depth,chen2025videodepthanything}, surface normal estimation~\cite{bin2025normalcrafter}, and semantic segmentation~\cite{videokmax,jain2019accel,miao2021vspw}.
%
%
Naturally, extending dense prediction to the video domain introduces additional annotation scarcity challenges.
This limitation results in existing unified video prediction methods being restricted to the self-driving domain~\cite{vtdnet}, primarily oriented toward video generation~\cite{udpdiff}, or confined to low-level geometric tasks~\cite{l4p}.
In contrast, \ours is not restricted to any domain, handles high-level semantic and intrinsic tasks beyond low-level geometry, and achieves temporal consistency through a lightweight parameter-free module without requiring task-specific video annotations.
\def\x{\mathbf{x}}
\def\y{\mathbf{y}}
\def\z{\mathbf{z}}
\def\l{\mathbf{l}}
\def\Rhw#1{\R^{h \times w \times #1}}
\def\RHW#1{\R^{H \times W \times #1}}
\def\Rthw#1{\R^{t \times h \times w \times #1}}
\def\RTHW#1{\R^{T \times H \times W \times #1}}
\def\Data{\mathbf{D}}
\def\Enc{\mathcal{E}}
\def\Dec{\mathcal{D}}
\def\Loss{\mathcal{L}}
\def\F{\mathcal{F}_\theta}  
\def\PixProj{\mathcal{P}}   
\def\G{\mathcal{G}_\phi}  
\def\LatProj{\Psi}   

\section{Method}
\label{sec:method}

\subsubsection{Task Formulation.}
\label{sec:task_form}
We consider the problem of dense scene understanding for videos, where a desired system must simultaneously predict $K$ tasks from an input video stream.
Formally, given an input video $v =\{\x_1, ..., \x_T\}$ with $T$ input frames $\x_i \in \RHW3$, the goal is to produce dense prediction maps $\{\y^k_1, ..., \y^k_T\}_{k=1}^K$, where $\y^k_i \in \RHW{C^k}$ is $i^\text{th}$-frame prediction for task $k$. 
Depending on the task, the output dimension $C^k$ can range from $1$ for scalar depth maps to the number of classes for semantic segmentation.
Furthermore, each task $k$ is associated with annotated data $\Data^k$ that may be completely disjoint.

%
We first learn task-specific embedding spaces that generalize beyond each task's limited training domain and modality (\Secref{sec:task_embed}), and then distill all task embeddings into a single unified video model (\Secref{sec:unified}).

\subsection{Learning Task-Specific Embeddings}
\label{sec:task_embed}

Unifying a diverse set of tasks often requires learning from disjoint data, as different tasks often require different means of label collection (\eg indoor RGB-D scans for geometric tasks versus manual annotations of internet images for semantic tasks).
To bridge these domain gaps, we leverage the rich generative priors of diffusion models pretrained on internet-scale data~\cite{rombach2022high}.

\subsubsection{Dense prediction with diffusion models.}
We build on a pretrained latent diffusion model (LDM)~\cite{rombach2022high} with a VAE encoder $\Enc$, decoder $\Dec$, and a U-Net backbone $\F$.
Given an input image $\x_i \in \RHW3$, the encoder maps it to a compact latent code $\z_i =\Enc(\x_i) \in \Rhw{d}$, while the decoder attempts to reconstruct the image $\x_i$ from $\z_i$ as $\Dec(\z_i) \in \RHW3$.

A key design choice is how to interface the network with the task output space.
%
%
%
We adopt a \emph{pixel decoding} strategy: the U-Net predicts a task latent $\l_i = \F(\z_i) \in \Rhw{d}$, which is decoded by the pretrained VAE decoder $\Dec$ up to its final convolutional layer, which we denote as $\Dec'(l_i) \in \RHW{d'}$.
We then project $\Dec'(l_i)$ to the target dimension $C^k$ of task $k$ via a lightweight task-specific pixel projector $\PixProj$.

For each task $k$, the final prediction $\hat{\y}$ is computed as follows:
\begin{equation}
    \l^k_i = \F^k\!\left(\Enc\!\left(\x_i\right)\right) \in \Rhw{d}, \quad 
    \hat{\y}^k_i = \PixProj^k\!\left(\Dec'\!\left(\l^k_i\right)\right) \in \RHW{C^k}
\end{equation}
where the task latent $\l^k_i$ is decoded by $\PixProj(\Dec'(\cdot))$ into $\hat{\y}^k_i \in \RHW{C^k}$.

The pixel decoding strategy (i) supports any $C^k$-dimensional outputs of arbitrary task, (ii) enables task-specific optimizations on the pixel predictions, and (iii) implicitly learns the projection between the latent and prediction space.

\subsubsection{Video predictions with temporal connections.}
Extending unified dense prediction to the video domain remains challenging due to annotation asymmetry across tasks (RGB-D videos for geometric tasks is more accessible than manual annotation for semantic or intrinsic tasks).
Therefore, a temporal mechanism should remain compatible for tasks with no temporal annotations.

We adopt extended self-attention~\cite{streamv2v}, which extends every self-attention layer in the U-Net backbone $\F$ with features from past frames.
At frame $t$, the queries attend jointly to the keys and values from both the current frame and a set of $M$ historical frames $\mathcal{M}_t \subseteq \{1, \ldots, t-1\}$ stored in a memory bank:
\begin{equation}
    \operatorname{Attn}\!\big(\mathbf{Q}_t,\, \left[\mathbf{K}_s\right]_{s \in \mathcal{M}_t \cup \{t\}},\, \left[\mathbf{V}_s\right]_{s \in \mathcal{M}_t \cup \{t\}}\big),
\end{equation}
where $[\cdot]$ denotes concatenation.
Intuitively, even without training on videos, adjacent frames would share rich correspondences~\cite{tang2023emergent} that can  encourage temporal smoothing as queries from frame $t$ attend to historical keys and values.

Setting $M = 0$ naturally reverts to single-frame predictions, making it compatible when training without temporal annotations.
The memory queue length $M$ and the enqueue/dequeue frequency can be chosen at inference, allowing flexibility to the degree of temporal smoothing and memory usage without retraining.

Now, an input video $v = \{ \x_1, ..., \x_T\}$ is projected to a task-specific embedding $\l = \left[ \F(\z_1), ..., \F(\z_T) \right] \in \R^{T \times h \times w \times d}$ via an $\F$ processing each $\z_i = \Enc(\x_i)$ with temporal connections, yielding a temporally-aware representation. 
%

\begin{figure}[t!]
  \centering
  \includegraphics[width=\textwidth]{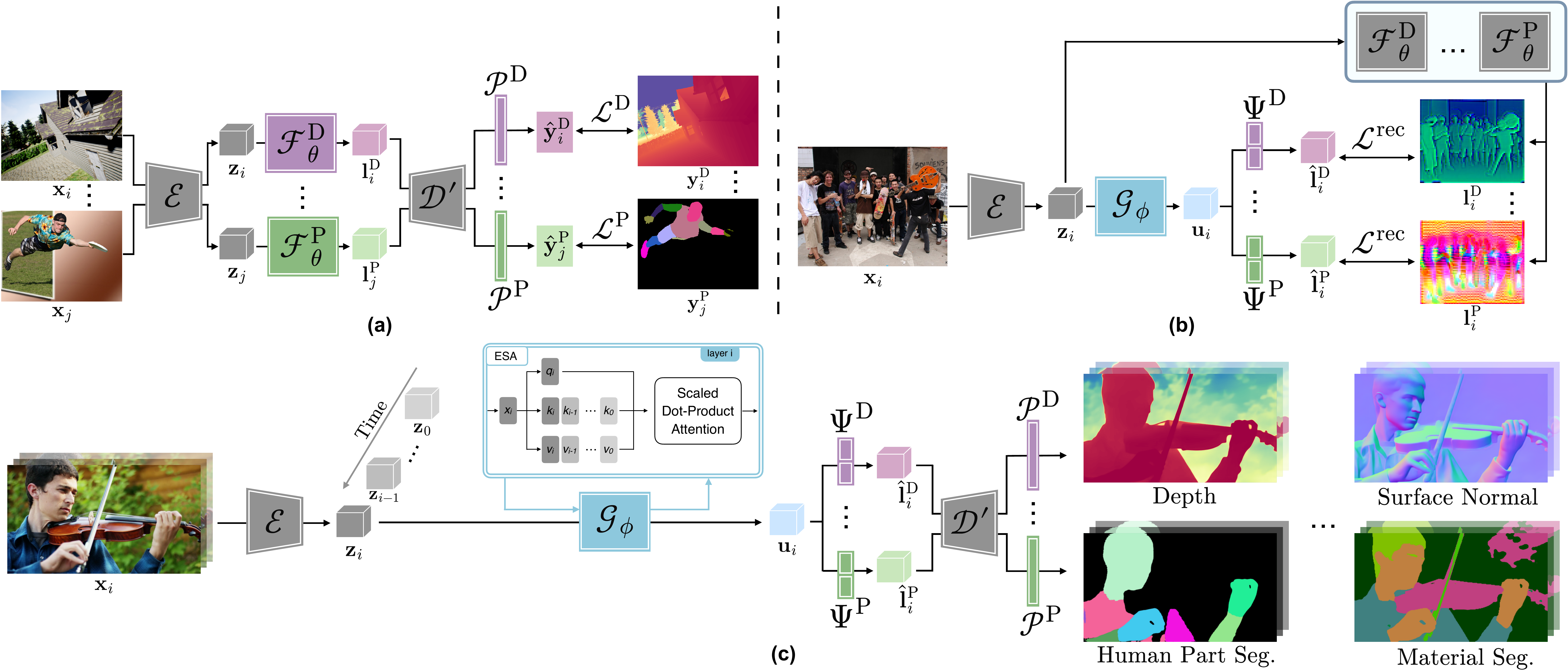}
  \caption{
        Overview of \ours. 
        \textbf{(a) Task specialist training.} For each task $k$, a specialist backbone $\F^k$ and pixel projector $\PixProj^k$ are finetuned on their task-specific dataset $\Data^k$ with loss $\mathcal{L}^k$. Note that both the encoder $\Enc$ and decoder $\Dec'$ (in gray) are frozen during training.
        \textbf{(b) Unified model training.} The unified backbone $\G$ is then trained along with $K$ latent projectors $\{\LatProj^k\}_{k=1}^K$ to reconstruct the latents $\{\l^k_i\}$ predicted by the frozen specialist $\{\F^k\}_{k=1}^K$ (in gray).
        \textbf{(c) Streaming video inference.} During inference, $\G$ processes an input video via Extended Self-Attention (ESA), then the unified representation $\mathbf{u}_i$ is decoded through each $\LatProj^k$, $\Dec'$, and $\PixProj^k$ to simultaneously produce temporally consistent predictions across all $K$ tasks.\vspace{-0.07cm}
}
  \label{fig:pipe}
\end{figure}

\subsubsection{Training task specialists.}
As shown in Figure~\ref{fig:pipe}(a), we train a dedicated specialist $(\F^k, \PixProj^k)$ for each task $k$ to learn a task-specific embedding that generalizes across disjoint, domain-specific data.
In line with the pixel-decoding framework, we also adopt a \emph{single-step} diffusion: rather than iterative denoising, the U-Net performs one forward pass to directly predict the latent $\l^k_i$.
The single-step diffusion not only reduces latency during inference, but also improves training efficiency by simplifying the original denoising objective~\cite{he2024lotus}.

Formally, for each task $k$, we obtain the task-specialist $(\F^k, \PixProj^k)$ by minimizing the task-specific objective $\Loss^k$ over the task-specific dataset $\Data^k$:
\begin{equation}
    \left(\F^k,\, \PixProj^k\right) \;=\; \argmin_{\F^k,\, \PixProj^k} \; \mathbb{E}_{(\x, \y) \sim \Data^k} \;\Big[ \Loss^k\!\Big(\PixProj^k\!\left(\Dec'\!\left(\F^k\!\left(\Enc(\x)\right)\right)\right),\, \y\Big) \Big]
\end{equation}
Due to its large-scale pretraining and generative priors, the learned $\F^k$ can bridge disjoint datasets when embedding unseen data during unified training.

\subsection{Unified Video Dense Prediction}
\label{sec:unified}

\subsubsection{Learning unified model with latent projectors.}
Given $K$ task specialists $\{(\F^k, \PixProj^k)\}_{k=1}^K$ in hand, one option would be to pseudo-label a large shared dataset with each specialist and train on the computed labels.
While feasible, computing/storing $K$ full-resolution label maps $\{\RHW{C^k}\}_{k=1}^K$ per frame is time-consuming and memory-intensive, with both costs growing linearly with $K$—all before unified training even begins.

Instead, we distill task embeddings directly in the latent space.
This allows us to avoid pseudo-labeling entirely, skip gradient propagation through the decoder $\Dec$, and unify the learning objective for each task as latent reconstruction.

For an unlabeled video $\x = \{\x_1, \ldots, \x_T\}$, we first compute the target task latents $\l^k_i$ with temporal connections on-the-fly via each frozen specialist backbone $\F^k$, where $\l^k_i \;=\; \F^k\!\left(\z_i\right) \in \Rhw{d}, \; \z_i = \Enc(\x_i), \; i = 1,...,T$.
Then, we formulate the unified model as a U-Net backbone $\G$, with same initialization and temporal memory as $\F^k$, and employ $K$ lightweight task-specific latent projectors $\{\LatProj^k\}_{k=1}^K$ to project the unified representation to the task-specific embeddings.
Given an input latent $\z_i$, the unified model predicts all task latents simultaneously:
\begin{equation}
    \hat{\l}^k_i \;=\; \LatProj^k\!\left(\G\!\left(\z_i\right)\right) \;\in\; \Rhw{d}, \quad k = 1, \ldots, K.
\end{equation}

As shown in Figure~\ref{fig:pipe}(b), the unified model $(\G, \{\LatProj^k\}_{k=1}^K)$ is learned by minimizing the latent reconstruction objective $\Loss^{\text{rec}}$ as follows:
\begin{equation}
    \left(\G,\, \{\LatProj^k\}\right) \;=\; \argmin_{\G,\, \{\LatProj^k\}} \; \mathbb{E}_{v \sim \Data} \; \sum_{k=1}^{K}  \Loss^{\text{rec}}\left(  \hat{\l}^k\;,\: \l^k \right)
\end{equation}
where $\hat{\l}^k$ and $\l^k$ are both computed by running $\LatProj^k(\G(\cdot))$ and $\F^k(\cdot)$ with temporal connections on the latent code $\z_i$ from $i=1$ to $i=T$.

\subsubsection{Temporal latent stabilization.}
If $\Loss^{\text{rec}}$ only contains per-frame reconstruction losses, the unified backbone $\G$ does not receive sufficient temporal regularization during unified training.
To address this, we extend temporal gradient matching~\cite{chen2025videodepthanything}, originally proposed for video depth estimation, to a general latent formulation.
Given predicted latents $\hat{\l}^k \in \Rthw{d}$ and target latents $\l^k \in \Rthw{d}$, we enforce that the temporal gradient of the prediction matches that of the target at every spatial location:
\begin{equation}
    \mathcal{L}_{\text{TGM}}(\hat{\l}, \l) \;=\; \left\| \big[ \,\big(\hat{\l}_{i+1} - \hat{\l}_i\big) - \big(\l_{i+1} - \l_i\big)\,\big]  \cdot \mathbb{1}\!\big[\left\|\l_{i+1} - \l_i\right\|_2 < \varepsilon\big] \right\|_2
\end{equation}
where the mask $\mathbb{1}[\cdot]$ suppresses learning at fast-changing regions where temporal regularization is not reliable.
The final unified training objective combines per-frame $\ell_1$ latent reconstruction with the temporal gradient matching loss: $\Loss^{\text{rec}}\left(  \hat{\l}^k\;,\: \l^k \right) = \left\|\ \hat{\l}^k_i - \l^k_i \ \right\|_1 + \lambda \, \mathcal{L}_{\text{TGM}}(\ \hat{\l}^k, \l^k\ )$, where $\lambda$ balances the two terms.

\subsubsection{Inference and extensibility.}
At inference, the full-resolution predictions for all tasks with input $\x \in \RTHW3$ are computed as follows:
\begin{equation}
    \mathbf{u}_i = \G\!\left( \Enc \left( \x_i \right) \right), \quad 
    \hat{\y}^k = \PixProj^k\!\left(\Dec'\!\left(\LatProj^k\!\left( \mathbf{u}_i \right)\right)\right) \in \RHW{C^k}, \quad 
    i = 1,...,T
\end{equation}
This framework is  naturally extensible: adding a new task $K{+}1$ requires only training a new specialist $(\F^{K+1}, \PixProj^{K+1})$ on its own labeled data and introducing a new latent projector $\LatProj^{K+1}$, with no architectural change to the unified backbone $\G$ and no existing data to be re-labeled.

\subsection{Implementation Details}
\label{sec:impl}



\paragraph{Task specialists.}
We use Stable Diffusion v2~\cite{rombach2022high} as our pretrained LDM backbone.
Each of $K = 8$ specialists finetunes the U-Net backbone along with the task-specific pixel projector $\PixProj^k$ (a single $3 \times 3$ convolution) to convergence with a learning rate of $3 \times 10^{-5}$ using the AdamW optimizer~\cite{loshchilov2017decoupled} with decay $10^{-2}$.
%
%

\paragraph{Unified model.}
The unified backbone $\G$ is initialized from the same pretrained U-Net.
Each latent projector $\LatProj^k$ is a DPT head that fuses $3$ intermediate representations of $\G$ with the final output feature at a fusion dimension of $256$.
The unified backbone $\G$ and latent projectors $\LatProj^k$ are trained with a union of all labeled data for all tasks, sampled with equal weight between image and video data.
The full unified model is trained for $50$k iterations with batch size of $32$ and learning rate $3 \times 10^{-5}$.
Please refer to the Appendix for details regarding each task's training datasets, learning objective, and other implementation details.

\section{Experiments}
\label{sec:experiments}

\subsubsection{Datasets and Metrics.}
The training data, evaluation data, task-specific output dimension $C^k$ and metrics are detailed in Table~\ref{tab:training_data}.
%

In terms of evaluation, we select at least one out-of-domain dataset (unseen during training) to evaluate the generalization of \ours.
For depth and surface normal, we follow the evaluation protocols of Marigold~\cite{marigold} and DSINE~\cite{dsine}, respectively.
For albedo and shading, we report Weighted Human Disagreement Rate (WHDR)~\cite{bell14intrinsic} and Precision at $70\%$ Recall ($P@0.7$)~\cite{kovacs17shading}, respectively.
For instance boundary, we report optimal-dataset-scale F-measure (odsF)~\cite{invpt}.
For segmentation tasks, we report meanIoU and class-frequency-weighted IoU (fwIoU).

To measure temporal consistency, we extend Video Consistency (mVC-t)~\cite{miao2021vspw} originally proposed for video semantic segmentation to also measure video normal, albedo, shading, and instance boundary.
%
%
Please refer to the appendix for details regarding evaluation benchmarks and metrics.

\subsubsection{Baselines.}
We compare \ours against per-task specialists and unified multi-task baselines.
Among the unified baselines, we include 4M-21-XL~\cite{bachmann20244m} and DICEPTION~\cite{diception}, both requiring large-scale pseudo-labeling.
Notably, DICEPTION~\cite{diception} requires task-specific captions as input, requiring $K$ forward passes to obtain all unified predictions.
We further include StableMTL~\cite{stablemtl}, which is trained only on synthetic data and likewise predicts tasks sequentially during inference.
As our most directly comparable baseline, we extend a frozen DINOv3-H~\cite{dinov3} backbone with task-specific DPT heads trained on exactly the same datasets with the same loss functions for 20k iterations.
For a fair comparison, the DINOv3-H backbone is chosen to have the same capacity as our unified backbone $\G$.
DINOv3-H simulates the alternative design of freezing a strong image representation and learning per-task decoders from disjoint sources, without any task-specific joint training or temporal modules.

\subsection{Per-Task Performance}
\label{sec:pertask}

\begin{figure}[t!]
  \centering
  \includegraphics[width=\textwidth]{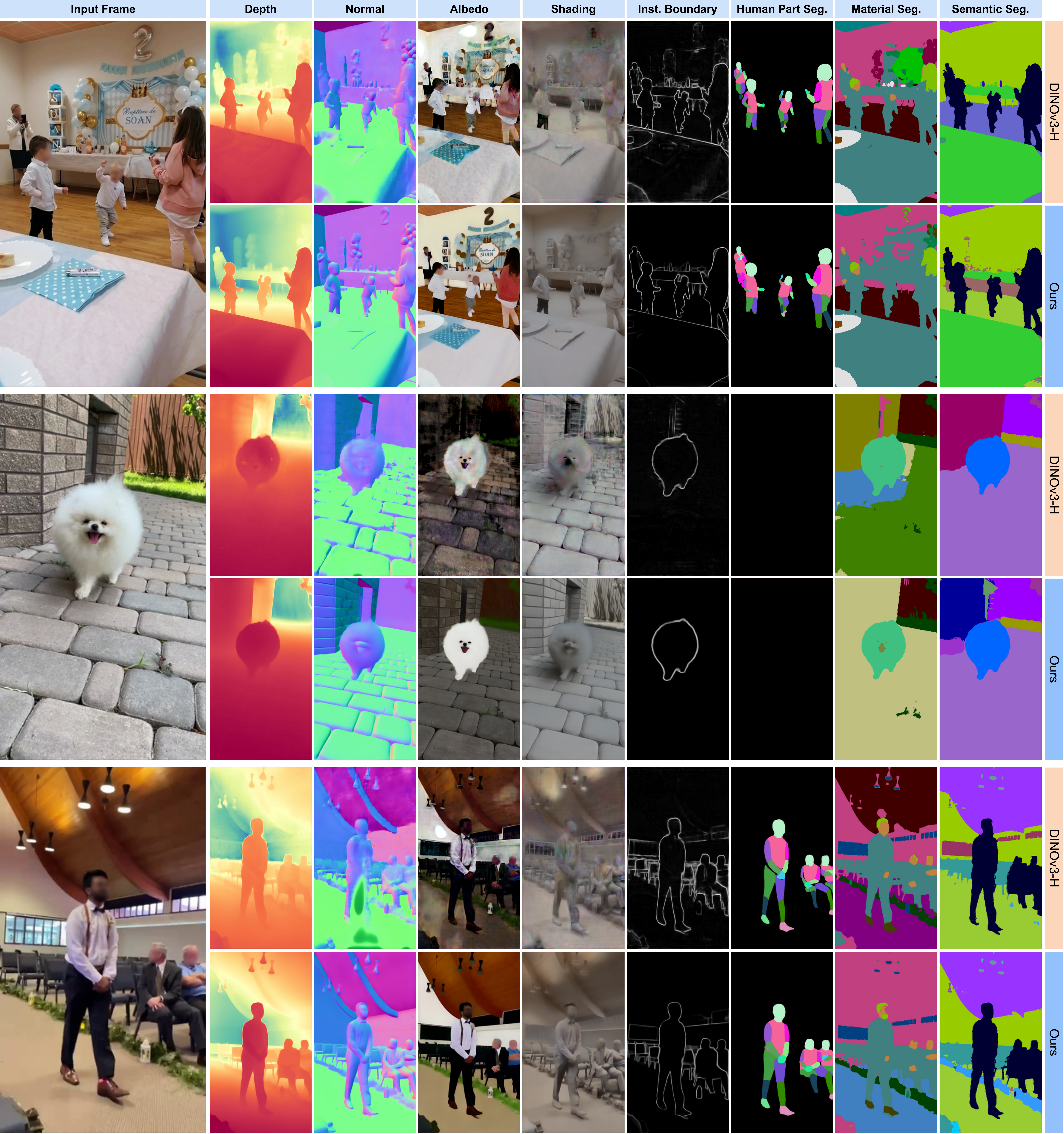}
  \caption{Qualitative Results of \ours compared with DINOv3-H.\protect\footnotemark}
  \label{fig:qual}
\end{figure}

\footnotetext{Visual examples taken from SA-V \cite{sam2} and internet videos \cite{pexels}.}

\begin{table}[t]
\centering
\caption{Summary of per-task output dimension $C^k$, training datasets (with frame counts), evaluation datasets, and metrics. Most training datasets are disjoint across task domains, with in-domain evaluation datasets shown in \setlength{\fboxsep}{2pt}\colorbox{gray!20}{gray}.}
\resizebox{\textwidth}{!}{%
\begin{tabular}{l@{\hspace{6pt}}c@{\hspace{6pt}}l@{\hspace{6pt}}r@{\hspace{9pt}}l@{\hspace{6pt}}l}
\toprule
\textbf{Task} & \textbf{$C^k$} & \textbf{Training Datasets} & \textbf{\#Frames} & \textbf{Evaluation Datasets} & \textbf{Metrics} \\
\midrule
\multirow{3}{*}{Depth}
  & \multirow{3}{*}{$1$}
  & \emph{Image}: Hypersim~\cite{hypersim}, VKITTI2~\cite{vkitti2} & $80.8$k
  & \multirow{2}{*}{NYUv2~\cite{nyuv2}, KITTI~\cite{kitti}, ETH3D~\cite{eth3d},}
  & \multirow{2}{*}{AbsRel$\downarrow$, $\delta_1\uparrow$} \\
  & & \emph{Video}: PointOdyssey~\cite{pointodyssey}, Spring~\cite{spring}, & \multirow{2}{*}{$361.0$k} & \multirow{2}{*}{ScanNet~\cite{scannet}, DIODE~\cite{diode}} & \\
  & & \hspace{0.92cm} TartanAir~\cite{tartanair} &  & & \\
\midrule
Surface Normals
  & $3$
  & \emph{Image}: Hypersim~\cite{hypersim}, VKITTI2~\cite{vkitti2} & $80.8$k
  & NYUv2~\cite{nyuv2}, ScanNet~\cite{scannet}, iBims-1~\cite{ibims}, Sintel~\cite{sintel}
  & Mean$\downarrow$, $11.25^\circ\uparrow$ \\
\midrule
Albedo
  & $3$
  & \emph{Image}: Hypersim~\cite{hypersim} & $59.5$k
  & \setlength{\fboxsep}{2pt}\colorbox{gray!20}{Hypersim~\cite{hypersim}}, IIW~\cite{bell14intrinsic}
  & PSNR$\uparrow$, WHDR$_{15\%}\downarrow$ \\
\midrule
Shading
  & $3$
  & \emph{Image}: Hypersim~\cite{hypersim} & $59.5$k
  & \setlength{\fboxsep}{2pt}\colorbox{gray!20}{Hypersim~\cite{hypersim}}, SAW~\cite{kovacs17shading}
  & PSNR$\uparrow$, P@0.7/AP$\uparrow$ \\
\midrule
\multirow{2}{*}{Instance Boundary}
  & \multirow{2}{*}{$1$}
  & \emph{Image}: COCO~\cite{coco}, Cityscapes~\cite{cityscapes} & $121.3$k
  & \multirow{2}{*}{\setlength{\fboxsep}{2pt}\colorbox{gray!20}{COCO~\cite{coco}}, YT-VIS~\cite{ytvis}, Mapillary~\cite{mapillary}}
  & \multirow{2}{*}{odsF$\uparrow$} \\
  & & \emph{Video}: VIPSeg~\cite{vipseg} & $66.8$k & & \\
\midrule
\multirow{2}{*}{Semantic Seg.}
  & $201/35/125$
  & \emph{Image}: COCO~\cite{coco}, Cityscapes~\cite{cityscapes} & $121.3$k
  & \multirow{2}{*}{\setlength{\fboxsep}{2pt}\colorbox{gray!20}{Cityscapes~\cite{cityscapes}}, \setlength{\fboxsep}{2pt}\colorbox{gray!20}{VIPSeg~\cite{vipseg}}, ADE20K~\cite{ade20k}}
  & \multirow{2}{*}{mIoU$\uparrow$, fwIoU$\uparrow$} \\
  & {\small(C/CS/VS)}
  & \emph{Video}: VIPSeg~\cite{vipseg} & $66.8$k & & \\
\midrule
Material Seg.
  & $57$
  & \emph{Image}: DMS~\cite{dms} & $22.0$k
  & \setlength{\fboxsep}{2pt}\colorbox{gray!20}{DMS~\cite{dms}}, ADE20K~\cite{ade20k}
  & mIoU$\uparrow$, fwIoU$\uparrow$ \\
\midrule
Human Part Seg.
  & $16$
  & \emph{Image}: DensePose~\cite{densepose} & $26.4$k
  & \setlength{\fboxsep}{2pt}\colorbox{gray!20}{DensePose~\cite{densepose}}, Pascal-Human-Part~\cite{pascalhuman}
  & mIoU$\uparrow$, fwIoU$\uparrow$ \\
\bottomrule
\end{tabular}
}
\label{tab:training_data}
\end{table}

\paragraph{Out-of-distribution generalization.}
The main claim of \ours is that the generative priors of a pretrained diffusion backbone are sufficient to bridge the domain gaps introduced by disjoint training sources, enabling generalization to scene-task combinations never seen with annotations.
To evaluate this, in-distribution benchmarks are shown in {\setlength{\fboxsep}{2pt}\colorbox{gray!20}{\color{darkgray!70}gray} in Tables~\ref{tab:intrinsic} and \ref{tab:seg}, and all remaining columns are out-of-distribution evaluations where labeled training data is not seen.

Across all geometric and intrinsics tasks, \ours ($\G$) consistently outperforms DINOv3-H on OOD benchmarks, despite identical training datasets.
For depth (Table~\ref{tab:depth}), \ours ($\G$) achieves the best AbsRel among all unified methods on NYUv2, ScanNet, and DIODE, while DINOv3-H underperforms by a consistent margin across all five datasets.
For surface normals (Table~\ref{tab:normal}), \ours ($\G$) achieves state-of-the-art performance among unified methods on all but one OOD benchmark metric.
For intrinsics and boundaries (Table~\ref{tab:intrinsic}), \ours ($\G$) matches DINOv3-H in-domain, while substantially outperforming it on every OOD benchmark: improving albedo WHDR from 0.289 to 0.207, shading P@0.7 from 87.8 to 94.6, and boundary odsF from 55.0 to 57.2.

This trend of parity on in-domain and clear advantage out-of-domain verifies the efficacy of the diffusion prior in transferring robustly to unseen scenarios.
To isolate the effect of diffusion prior, we apply color jittering to Hypersim val~\cite{hypersim} and found the specialist \ours ($\F^k$) to have substantially more stable predictions than DINOv3-H on depth (AbsRel deviation: 0.011 vs.\ 0.034) and surface normals (mean error deviation: 2.2° vs.\ 5.8°) --- roughly $3\times$ \textbf{more robust} despite identical training data.
This suggests that discriminative features are prone to overfit to the appearance-geometry correlations in the training data, while generative pretraining on web-scale data encodes more appearance-invariant geometry.
This is further evident in Figure~\ref{fig:qual}, where DINOv3-H predictions degrade noticeably for unseen scenes while \ours ($\G$) remains coherent across all tasks.

\begin{table}[t]
\centering
\caption{Monocular Depth Estimation Performance on NYUv2, KITTI, ETH3D, ScanNet, and DIODE. Best performance among unified methods is bolded.}
\resizebox{\textwidth}{!}{%
\begin{tabular}{@{\hspace{9pt}}c@{\hspace{9pt}}l cc@{\hspace{9pt}}cc@{\hspace{9pt}}cc@{\hspace{9pt}}cc@{\hspace{9pt}}cc}
\toprule
\multirow{2}{*}{} & \multirow{2}{*}{Method} & \multicolumn{2}{c}{NYUv2} & \multicolumn{2}{c}{KITTI} & \multicolumn{2}{c}{ETH3D} & \multicolumn{2}{c}{ScanNet} & \multicolumn{2}{c}{DIODE} \\
\pad \cline{3-12} \pad
 & &  $\text{AbsRel} \downarrow$ & $\delta_1$$\uparrow$ & $\text{AbsRel} \downarrow$ & $\delta_1$$\uparrow$ & $\text{AbsRel} \downarrow$ & $\delta_1$$\uparrow$ & $\text{AbsRel} \downarrow$ & $\delta_1$$\uparrow$ & $\text{AbsRel} \downarrow$ & $\delta_1$$\uparrow$ \\
\midrule
\multirow{7}{*}{\rotatebox[origin=c]{90}{Specialist}}
& DAv2~\cite{depthanythingv2}			& 0.043	& 98.1	& 0.076	& 94.7	& 0.127	& 88.2	& 0.043	& 98.1	& 0.260	& 75.9	\\
& Depth Pro~\cite{bochkovskii2024depthpro}			& 0.042	& 97.7	& 0.055	& 97.4	& 0.043	& 97.4	& 0.041	& 97.8	& 0.217	& 76.4	\\
& Marigold~\cite{marigold}			& 0.055	& 96.4	& 0.099	& 91.6	& 0.065	& 95.9	& 0.064	& 95.2	& 0.308	& 77.3	\\
& Genpercept~\cite{xu2024matters}			& 0.091	& 93.2	& 0.094	& 92.3	& 0.066	& 95.7	& 0.056	& 96.5	& 0.302	& 76.7	\\
& GeoWizard~\cite{fu2024geowizard}			& 0.052	& 96.6	& 0.097	& 92.1	& 0.064	& 96.1	& 0.061	& 95.3	& 0.297	& 79.2	\\
& Lotus~\cite{he2024lotus}			& 0.051	& 97.2	& 0.081	& 93.1	& 0.061	& 97.0	& 0.055	& 96.5	& 0.228	& 73.8	\\
& \ours ($\F^\text{ D}$)      & 0.059	& 96.0	& 0.102	& 90.9	& 0.070	& 95.4	& 0.069	& 95.0	& 0.299	& 77.4	\\
\midrule
\multirow{5}{*}{\rotatebox[origin=c]{90}{Unified}}
& 4M-21-XL~\cite{bachmann20244m}			& 0.068	& 95.1	& 0.105	& 89.6	& \textbf{0.070}	& 95.3	& 0.065	& 95.5	& 0.331	& 73.4	\\
& DICEPTION~\cite{diception}			& 0.060	& \textbf{96.3}	& \textbf{0.085}	& \textbf{91.8}	& \textbf{0.070}	& \textbf{96.6}	& 0.071	& 94.8	& 0.304	& 72.2	\\
& StableMTL~\cite{stablemtl} & 0.090 & 92.4 & 0.150 & 79.8 & 0.116 & 87.9 & 0.105 & 89.5 & 0.330 & 73.8\\
& DINOv3-H~\cite{dinov3}			& 0.092	& 92.9	& 0.121	& 86.5	& 0.087	& 93.7	& 0.089	& 93.6	& 0.314	& \textbf{77.5}	\\
 & \ours ($\G$)           & \textbf{0.059}	& \textbf{96.3}	& 0.110	& 89.5	& 0.073	& 95.5	& \textbf{0.063}	& \textbf{96.2}	& \textbf{0.301}	& 77.3 \\
\bottomrule
\end{tabular}
}
\label{tab:depth}
\end{table}
\begin{table}[t]
\centering
\caption{Surface Normal Estimation Performance on NYUv2, ScanNet, iBims-1, and Sintel. Best performance among unified methods is bolded.}
\resizebox{0.9\textwidth}{!}{%
\begin{tabular}{@{\hspace{9pt}}c@{\hspace{9pt}}l cc@{\hspace{9pt}}cc@{\hspace{9pt}}cc@{\hspace{9pt}}cc}
\toprule
\multirow{2}{*}{} & \multirow{2}{*}{Method} & \multicolumn{2}{c}{NYUv2} & \multicolumn{2}{c}{ScanNet} & \multicolumn{2}{c}{iBims-1} & \multicolumn{2}{c}{Sintel} \\
\pad \cline{3-10} \pad
 & &  $\text{mean} \downarrow$ & $11.25^\circ$$\uparrow$ & $\text{mean} \downarrow$ & $11.25^\circ$$\uparrow$ & $\text{mean} \downarrow$ & $11.25^\circ$$\uparrow$ & $\text{mean} \downarrow$ & $11.25^\circ$$\uparrow$ \\
\midrule
\multirow{7}{*}{\rotatebox[origin=c]{90}{Specialist}}
& OASIS~\cite{chen2020oasis}			& 29.2	& 23.8	& 32.8	& 15.4	& 32.6	& 23.5	& 43.1	& 7.0	\\
& Omnidata v2~\cite{kar20223d}			& 17.2	& 55.5	& 16.2	& 60.2	& 18.2	& 63.9	& 40.5	& 14.7	\\
& DSINE~\cite{dsine}			& 16.4	& 59.6	& 16.2	& 61.0	& 17.1	& 67.4	& 34.9	& 21.5	\\
& GeoWizard~\cite{fu2024geowizard}			& 18.9	& 50.7	& 17.4	& 53.8	& 19.3	& 63.0	& 40.3	& 12.3	\\
& StableNormal~\cite{ye2024stablenormal}			& 18.6	& 53.5	& 17.1	& 57.4	& 18.2	& 65.0	& 36.7	& 14.1	\\
& Lotus~\cite{he2024lotus}			& 16.2	& 59.8	& 14.7	& 64.0	& 17.1	& 66.4	& 32.3	& 22.4	\\
& \ours ($\F^\text{ N}$)			& 16.7	& 59.9	& 15.4	& 63.7	& 16.3	& 69.0	& 34.3	& 22.2	\\
\midrule
\multirow{5}{*}{\rotatebox[origin=c]{90}{Unified}}
& 4M-21-XL~\cite{bachmann20244m}			& 18.4	& 49.8	& 17.4	& 48.9	& 18.6	& 61.4	& 41.6	& 12.4	\\
& DICEPTION~\cite{diception}			& 19.7	& 51.4	& 19.3	& 54.5	& 17.8	& 66.2	& 37.2	& 19.6	\\
& StableMTL~\cite{stablemtl}  & 19.6 & 48.9 & 20.0 & 45.8 & 19.6 & 58.1 & 40.5 & 11.0\\
& DINOv3-H~\cite{dinov3}			& 16.8	& 57.0	& 16.0	& 58.7	& 16.7	& 66.8	& \textbf{32.3}	& 18.9	\\
& \ours ($\G$) 			& \textbf{16.2}	& \textbf{59.9}	& \textbf{14.6}	& \textbf{65.1}	& \textbf{16.6}	& \textbf{67.4}	& 33.0	& \textbf{21.4}	\\

\bottomrule
\end{tabular}
}
\label{tab:normal}
\end{table}
\begin{table}[t]
\centering
\caption{Albedo, Shading, and Instance Boundary Estimation Performance. Best performance among unified methods is bolded.}
\resizebox{\textwidth}{!}{%
\begin{tabular}{@{\hspace{9pt}}c@{\hspace{9pt}}l cc@{\hspace{12pt}}cc@{\hspace{12pt}}cc}
\toprule
\multirow{3}{*}{} & \multirow{3}{*}{Method} & \multicolumn{2}{c}{\textbf{Albedo}} & \multicolumn{2}{c}{\textbf{Shading}} & \multicolumn{2}{c}{\textbf{Boundary}}\\
&  & \gcellonly Hypersim & \multicolumn{1}{c}{IIW} & \gcellonly Hypersim & \multicolumn{1}{c}{SAW} & \gcellonly COCO	 & Mapillary\\
\pad \cline{3-8} \pad
 & &  \gcellonly$\text{PSNR} \uparrow$ & $\text{WHDR}_{15\%} \downarrow$  & \gcellonly$\text{PSNR} \uparrow$ & $\text{P}@0.7 \text{/AP}\% \uparrow$ & \gcellonly$\text{odsF} \uparrow$ & $\text{odsF} \uparrow$ \\
\midrule
\multirow{3}{*}{\rotatebox[origin=c]{90}{Spec.}}
& ColorfulShading~\cite{careaga2024colorful} &\gcell17.2 & 0.175 & \gcell14.7  & 81.3 / 79.4   & \gcell- & -  \\
& Mask2Former~\cite{mask2former} & \gcell-		& -	& \gcell-		& -  &\gcell67.7 & 35.6 \\
& \ours ($\F^k$)        & \gcell17.8		& 0.201	& \gcell20.3		& 93.5	/ 91.5	& \gcell79.8			& 56.3	\\
\midrule
\multirow{3}{*}{\rotatebox[origin=c]{90}{Unif.}}
& DINOv3-H~\cite{dinov3}                   & \gcell\textbf{17.8}		& 0.289	& \gcell19.5		& 87.8	/ 88.0	& \gcell\textbf{81.4}			& 55.0	\\
& StableMTL~\cite{stablemtl}  & \gcell - & \textbf{0.190} & \gcell - & 82.7  / 63.7 & \gcell - & -\\
& \ours ($\G$) 		                & \gcell17.6		& 0.207	& \gcell\textbf{20.2}		& \textbf{94.6	/ 92.6}	& \gcell80.8	& \textbf{57.2}	\\         

\bottomrule
\end{tabular}
}
\label{tab:intrinsic}
\end{table}

\begin{table}[t]
\centering
\caption{Semantic, Material, and Human Part Segmentation Performance.}
\resizebox{\textwidth}{!}{%
\begin{tabular}{@{\hspace{9pt}}c@{\hspace{9pt}}l ccc@{\hspace{12pt}}cc@{\hspace{12pt}}cc}
\toprule
\multirow{3}{*}{} & \multirow{3}{*}{Method} & \multicolumn{3}{c}{\textbf{Semantic}} & \multicolumn{2}{c}{\textbf{Material}} & \multicolumn{2}{c}{\textbf{Human Part}}\\
&  & \gcellonly Cityscapes & \gcellonly VIPSeg & ADE20K & \gcellonly DMS & ADE20K & \gcellonly DensePose &Pascal-Human-Part	\\
\pad \cline{3-9} \pad
 & & \gcellonly mIoU / fwIoU & \gcellonly mIoU / fwIoU & mIoU / fwIoU & \gcellonly mIoU / fwIoU & mIoU / fwIoU & \gcellonly mIoU / fwIoU & mIoU / fwIoU\\
\midrule
Spec.
& \ours ($\F^k$)        & \gcell64.1	/ 85.3	& \gcell45.8	/ 67.2	& 47.6	/ 71.6	& \gcell47.1	/ 74.1	& 67.2	/ 76.2	& \gcell68.2	/ 93.3	& 66.9	/ 88.4	\\
\midrule
\multirow{4}{*}{\rotatebox[origin=c]{90}{Unified}}
& StableMTL~\cite{stablemtl} & \gcell55.8 / -	& \gcell -/-	& -/-	& \gcell -/-	& -/-	& \gcell -/-	& -/-	\\
& DINOv3-H~\cite{dinov3}                    & \gcell65.6	/ 83.8	& \gcell52.8	/ 66.8	& 62.4	/ 74.8	& \gcell50.6	/ 75.1	& 69.1	/ 75.7	& \gcell67.7	/ 93.0	& 69.0	/ 89.1	\\
& \ours ($\G$) 		                 & \gcell37.8	/ 79.7	& \gcell46.1	/ 67.1	& 35.7	/ 66.2	& \gcell40.6	/ 72.1	& 44.1	/ 81.4	& \gcell64.3	/ 92.9	& 65.9	/ 88.3	\\      
& \ours ($\G$) + FT 		         & \gcell58.4	/ 83.3	& \gcell47.0	/ 67.0	& 44.8	/ 68.9	& \gcell47.3	/ 73.6	& 66.0	/ 76.0	& \gcell66.1	/ 92.9	& 65.4	/ 88.0	\\
\bottomrule
\end{tabular}
}
\label{tab:seg}
\end{table}

\begin{table}[t]
\centering
\caption{Temporal Consistency performance on video prediction tasks. Best performance among unified methods is bolded. Extended Self-Attention~\cite{streamv2v} is additionally added to image models with 16-frame memory dequeued at every frame (+ESA[16,1]).}
\resizebox{\textwidth}{!}{%
\begin{tabular}{@{\hspace{9pt}}c@{\hspace{9pt}}l cc@{\hspace{9pt}}c@{\hspace{9pt}}c@{\hspace{9pt}}c@{\hspace{9pt}}c@{\hspace{9pt}}c}
\toprule
\multirow{3}{*}{} & \multirow{3}{*}{Method} & \multicolumn{2}{c}{\textbf{Depth}} & \textbf{Normal} & \textbf{Albedo} & \textbf{Shading} & \textbf{Boundary} & \textbf{Semantic} \\
 &  & \multicolumn{2}{c}{ScanNet} & InteriorNet & InteriorNet & InteriorNet & YT-VIS & VIPSeg \\
\pad \cline{3-9} \pad
 & &  $\text{AbsRel} \downarrow$ & $\delta_1$$\uparrow$  & $\text{mVC-4/16} \uparrow$ & $\text{mVC-4/16} \uparrow$ & $\text{mVC-4/16} \uparrow$ & $\text{mVC-4} \uparrow$ & $\text{mVC-4} \uparrow$ \\
\midrule
\multirow{4}{*}{\rotatebox[origin=c]{90}{Spec.}}
& Video Depth Anything~\cite{chen2025videodepthanything}  & 0.063	& 96.8	 & -/- & -/- & -/- & - & - \\
& NormalCrafter~\cite{bin2025normalcrafter}         & - & -  & 84.6	/ 70.9 & -/- & -/- & - & - \\
& ColorfulShading~\cite{careaga2024colorful}                & -	& -	 & - / -	& 78.9 / 55.1	& 73.5 / 41.5	& -	& -	\\
& \ours ($\F^k$)                & 0.101	& 91.2	 & 85.9 / 76.4	& 86.9 / 70.1	& 92.5 / 81.9	& 93.6	& 93.8	\\
\midrule
\multirow{5}{*}{\rotatebox[origin=c]{90}{Unified}}
& DICEPTION~\cite{diception}             & 0.123	& 85.7		& 68.0 / 47.1	 & -/- & -/- & - & - \\
& DINOv3-H~\cite{dinov3}               & 0.120	& 86.9		& 78.2 / 62.2	& 77.8 / 53.4	& 90.2 / 77.6	& 92.0	& 87.8	\\
& DICEPTION~\cite{diception} + ESA[16,1]  & 0.118 & 85.9    & 66.0 / 48.6   & -/- & -/- & - & - \\
& DINOv3-H~\cite{dinov3} + ESA[16,1]   & 0.150 & 79.8 & 79.0 / 63.6 &  76.4 / 54.1 & 92.4 / 82.7 & 92.4 & 89.7 \\
& \ours ($\G$)                  & \textbf{0.103}	& \textbf{91.0}		& \textbf{84.7 / 75.8}	& \textbf{88.8 / 73.7}	& \textbf{94.5 / 85.2}	& \textbf{93.7}	& \textbf{90.3}	\\
\bottomrule
\end{tabular}
}
\label{tab:temp_const}
\end{table}

\paragraph{Unified model versus task-specialist.}
Across geometric and intrinsic tasks, \ours ($\G$) achieves similar performance with per-task specialists $\F^k$.
This demonstrates the efficacy of latent distillation that faithfully preserves specialist-level representation even after unifying all eight tasks into a single shared backbone.
For surface normals, the unified model $\G$ in fact \emph{outperforms} its own specialist $\F^\text{N}$ on NYUv2, ScanNet, and Sintel, a gain that can be attributed to cross-task regularization from depth, which shares the same geometric training domain and jointly reinforces surface geometry estimation within the unified backbone.
Furthermore, for shading and instance boundaries (Table~\ref{tab:intrinsic}), \ours ($\G$) matches the specialist $\F^k$ on in-domain evaluations, while improving it marginally on out-of-domain evaluations.

\paragraph{Segmentation and lightweight fine-tuning.}
In contrast, Table~\ref{tab:seg} reveals a systematic gap between \ours ($\G$) and the per-task specialist for classification tasks.
We find that reconstruction via $\ell_1$ is insufficient to fully match the specialists learned under cross-entropy (CE) instead of regression-based objectives, and this degradation persists even when $K{=}1$ (\ie when only distilling a single segmentation specialist), indicating the limitation is intrinsic to the latent reconstruction objective.
We trace this to a projector-backbone misalignment: CE-task specialist latents exhibit significantly higher spatial Laplacian norms (2.52 vs.\ 1.40) than regression-task latents, suggesting that $\ell_1$ distillation may suppress the high-frequency structure that CE specialists encode.
To address this, we perform a lightweight fine-tuning step that only updates the task-specific projectors $(\LatProj^k, \PixProj^k)$ while keeping $\G$ entirely frozen.
This targeted fine-tuning (reported as \ours ($\G$) + FT) substantially closes the gap with the specialist, recovering from 37.8 to 58.4 in mIoU for Cityscapes and from 35.7 to 44.8 in mIoU for ADE20K.
The rapid recovery from projector-only finetuning confirms that the backbone $\G$ \emph{does} learn semantically rich representations under $\ell_1$ distillation, preserving the unified representation across all tasks.

\subsection{Temporal Consistency}
\label{sec:temporal}

\begin{table}[t]
\centering
\caption{Cross-Task Consistency Evaluation. Best performance among unified methods is bolded.}
\resizebox{0.75\textwidth}{!}{%
\begin{tabular}{@{\hspace{9pt}}c@{\hspace{9pt}}l  cc@{\hspace{18pt}}c@{\hspace{18pt}}cc}
\toprule
\multirow{3}{*}{} & \multirow{3}{*}{Method} 
    & \multicolumn{2}{c}{\textbf{Depth$\leftrightarrow$Normal}}
    & \textbf{Part$\leftrightarrow$Semantic}
    & \multicolumn{2}{c}{\textbf{Albedo$\leftrightarrow$Shading}}\\
 & & \multicolumn{2}{c}{ScanNet} & COCO & \multicolumn{2}{c}{HyperSim}  \\
\pad \cline{3-7} \pad
 & & $\text{mean}\downarrow$ & $11.25^\circ\uparrow$ 
   & $\text{\% inclusion}\uparrow$ 
   & $\text{PSNR}\uparrow$ & $\text{LPIPS}\downarrow$ \\
\midrule
\multirow{2}{*}{\rotatebox[origin=c]{90}{Spec.}}
& ColorfulShading~\cite{careaga2024colorful} & -&- & - & 20.8 & 0.27 \\
& \ours ($\F^k$)  & 23.9 & 39.0	& 97.6	& 24.6	& 0.22 \\
\midrule
\multirow{4}{*}{\rotatebox[origin=c]{90}{Unified}}
& 4M-21-XL~\cite{bachmann20244m}			          & 23.4 & 37.2 & - &- &- \\
& DICEPTION~\cite{diception} & 27.3 & 27.4 & - & - &- \\
& DINOv3-H~\cite{dinov3} & 36.0 & 14.1	& 98.1	& 23.9 & 0.26  \\
& \ours ($\G$) & \textbf{22.6} & \textbf{39.0}	& \textbf{98.3}	& \textbf{24.3} & \textbf{0.24}  \\
\bottomrule
\end{tabular}
}
\label{tab:cross_task}
\end{table}

Table~\ref{tab:temp_const} evaluates temporal consistency across depth, normals, albedo, shading, boundary, and semantic segmentation on video benchmarks.
\ours ($\G$) outperforms all unified baselines across every task and, notably, is competitive with video-specific specialists trained for individual tasks.
On depth, our streaming model outperforms DICEPTION and DINOv3-H, while marginally underperforming Video Depth Anything that processes video in a chunk-based fashion.
For surface normals, \ours ($\G$) outperforms DINOv3-H by a large margin, and even surpasses NormalCrafter~\cite{bin2025normalcrafter} that utilizes a video diffusion backbone.

To verify that the advantage over unified baselines is not merely an artifact of the baselines lacking temporal module, we additionally equip DINOv3-H and DICEPTION with the same extended self-attention (ESA) for a fair comparison.
We find that ESA yields marginal but inconsistent improvements for both baselines, while \ours ($\G$) still outperforms all variants across all tasks.
This demonstrates that temporal stability arises not from the parameter-free ESA module alone, but from its combination with unified training on video data, which together propagate temporal smoothness across all tasks simultaneously without requiring task-specific video annotations.

\subsection{Cross-Task Consistency}
\label{sec:crosstask}

Table~\ref{tab:cross_task} evaluates geometric, semantic, and photometric consistency between pairs of related task predictions.

To measure consistency between depth and normal, we compute error between surface normal derived from depth predictions and direct normal prediction.
\ours ($\G$) achieves the best consistency on ScanNet among all methods, outperforming DICEPTION and DINOv3-H via a wide margin, and surpassing the independent specialist ensemble \ours ($\F^k$) .
For consistency between human part and semantic segmentation, we measure the \% inclusion of predicted human part pixels within the human semantic category.
Here, the unified model marginally outperforms DINOv3-H and the specialist ensemble, confirming that the shared backbone encourages semantic predictions to remain coherent with each other.
Finally, to measure intrinsic consistency, we directly measure reconstruction performance of diffuse reconstruction of the predicted albedo and shading.
Similarly, \ours ($\G$) remains competitive with the specialist ensemble, while still outperforming DINOv3-H.

This suggests that the joint latent distillation objective—where a unified backbone is supervised by all task specialists simultaneously—implicitly encourages cross-task geometric and semantic coherence that independent specialists or frozen-backbone baselines cannot achieve.

\begin{table}[t]
\centering
\caption{Per-component inference analysis on a single A100 GPU per-frame at $432{\times}768$ resolution with varying $M$ number of past frames in memory. UniD~($\F^k$) denotes the specialist ensemble ($K{=}8$ sequential U-Net passes); UniD~($\G$) denotes the unified model (one shared U-Net pass with $K$ DPT connectors $\LatProj^k$).}
\resizebox{\textwidth}{!}{%
\begin{tabular}{l@{\hspace{9pt}}c@{\hspace{18pt}}c@{\hspace{18pt}}c@{\hspace{18pt}}c@{\hspace{18pt}}c@{\hspace{18pt}}c@{\hspace{18pt}}c@{\hspace{9pt}}c}
\toprule
\textbf{Method} & $M$ & $\Enc$ (ms) & \textbf{U-Net} (ms) & $\LatProj^k$ (ms) & $\Dec$ (ms) & $\PixProj^k$ (ms) & \textbf{Total (s)}  & \textbf{FPS} \\
\midrule
StableMTL~\cite{stablemtl} & $0$ & - & - & - & - & - & $0.03 + 0.58 \times K$	& 0.21 \\
4M-21-XL~\cite{bachmann20244m} & $0$ & - & - & - & - & - & $0.29 + 0.43 \times K$	& 0.27 \\
DICEPTION~\cite{diception} & $0$ & - & - & - & - & - & $0.07 + 4.15 \times K$	& 0.03 \\
DINOv3-H~\cite{dinov3} & $0$ & - & - & - & - & - & $\mathbf{0.21 + 0.02 \times K}$	& \textbf{2.67} \\
\ours~($\F^k$) & $0$ & 35.9	& $58.6 \times K$	& -	& $42.3 \times K$	& $1.1 \times K$	& $0.04 + 0.10 \times K$	& 1.17 \\
\ours~($\G$) & $0$ & 35.9	& 58.6	& $3.1 \times K$	& $42.3 \times K$	& $1.1 \times K$	& $0.09 + 0.05 \times K$	& 2.14 \\
\midrule
DICEPTION~\cite{diception} + ESA & $16$ & - & - & - & - & - & $0.07 + 5.05 \times K$	& 0.02 \\
DINOv3-H~\cite{dinov3} + ESA & $16$ & - & - & - & - & - & $1.30 + 0.02 \times K$	& 0.69 \\
\ours~($\F^k$) & $16$ & 36.5	& $235.0 \times K$	& -	& $42.2 \times K$	& $1.1 \times K$	& $0.04 + 0.28 \times K$	& 0.44 \\
\ours~($\G$) & $16$ & 36.5	& 235.0	& $3.0 \times K$	& $42.2 \times K$	& $1.1 \times K$	& $\mathbf{0.27 + 0.05 \times K}$	& \textbf{1.56} \\
\bottomrule
\end{tabular}
}
\label{tab:time}
\end{table}
\begin{table}[t]
\centering
\caption{Ablation Study on temporal consistency performance on video prediction tasks. If \emph{Train w/ Video} is set to \xmark, all videos are treated as independent images.}
\resizebox{\textwidth}{!}{%
\begin{tabular}{cc @{\hspace{12pt}}  cc@{\hspace{9pt}}c@{\hspace{9pt}}c@{\hspace{9pt}}c@{\hspace{9pt}}c@{\hspace{9pt}}c}
\toprule
\multirow{3}{*}{} &  & \multicolumn{2}{c}{\textbf{Depth}} & \textbf{Normal} & \textbf{Albedo} & \textbf{Shading} & \textbf{Boundary} & \textbf{Semantic} \\
Train w/ & Inf. & \multicolumn{2}{c}{ScanNet} & InteriorNet & InteriorNet & InteriorNet & YT-VIS & VIPSeg \\
\pad \cline{3-9} \pad
 Video & Memory &  $\text{AbsRel} \downarrow$ & $\delta_1$$\uparrow$  & $\text{mVC-4/16} \uparrow$ & $\text{mVC-4/16} \uparrow$ & $\text{mVC-4/16} \uparrow$ & $\text{mVC-4} \uparrow$ & $\text{mVC-4} \uparrow$ \\
\midrule
\xmark  & $\emptyset$    & 0.121	& 87.1		& 79.0 / 65.6	& 84.4 / 65.0	& 92.7 / 81.5	& 92.9	& 85.9	\\
\xmark  & $[16,1]$       & 0.115	& 88.1		& 82.9 / 72.7	& 87.3 / 70.7	& 93.6 / 82.9	& 93.0	& 89.1	\\
\cmark  & $\emptyset$    & 0.109	& 90.0		& 80.9 / 68.5	& 86.0 / 67.1	& 93.7 / 83.5 & 93.6	& 86.5\\
\midrule
\cmark  & $[16,1]$       & 0.103	& 91.0		& 84.7 / 75.8	& 88.8 / 73.7	& 94.5 / 85.2	& 93.7	& 90.3	\\
\bottomrule
\end{tabular}
}
\label{tab:temp_abl}
\end{table}

\subsection{Efficiency Analysis \& Ablation Studies}

\subsubsection{Efficiency Analysis.}
In Table~\ref{tab:time}, \ours~($\G$) is efficient across all memory settings, remaining competitive with the discriminative DINOv3-H~\cite{dinov3} at $M{=}0$ while substantially outperforming all baselines at $M{=}16$.
StableMTL~\cite{stablemtl}, DICEPTION~\cite{diception}, and 4M-21-XL~\cite{bachmann20244m} underperform due to their prohibitive per-task costs, whereas \ours~($\G$) sustains $1.56$ FPS even with a $16$-frame memory bank.

The key to this efficiency is that nearly all computation is shared across tasks.
Compared to the specialist ensemble \ours~($\F^k$), which requires $K$ sequential U-Net passes with memory attentions, \ours~($\G$) processes a single \emph{unified} historical feature for all tasks at once, yielding a $1.8\times$ speedup at $M{=}0$ that compounds to over $\mathbf{3.5\times}$ at $M{=}16$ as temporal context grows.

\subsubsection{Latent versus per-pixel distillation.}
A natural alternative to our proposed latent distillation would be to directly supervise the unified model with per-pixel predictions with either pre-computed pseudo-labels or on-the-fly.
However, they are infeasible at scale due to the prohibitive nature of per-pixel backpropagation during unified training.
Computing gradients through the decoder $\Dec'$ per task in a training step requires holding $K$ separate decoder activation graphs in GPU memory simultaneously, growing memory consumption linearly with $K$ and rendering joint training intractable on standard hardware.
Latent distillation sidesteps this bottleneck: latent targets $\l^k_i = \F^k(\z_i) \in \Rhw{d}$ are cheap to compute, small to store, and avoid any computation with the decoder $\Dec$.
As a result, with identical training setup, the peak memory usage per device can decrease from an estimated $650$GB to $78$GB, when switching to latent distillation.

\subsubsection{Impact of video training and inference-time memory.}
Shown in Table~\ref{tab:temp_abl}, when all videos are treated as independent images during training, temporal consistency degrades across all benchmarks.
This directly ablates the temporal gradient matching (TGM) loss, as static-image training removes all cross-frame supervision. 
This suggests that, with only depth and semantic segmentation containing video annotations, unified training propagates temporal consistency to all tasks, even those without task-specific video annotations.

Notably, this holds even with no inference memory ($\emptyset$): video training alone yields consistently more stable representations across all tasks, before any inference-time memory module is introduced.
Enabling extended self-attention memory bank at inference ($[16, 1]$, a queue of 16 frames dequeued every frame) further improves temporal consistency across all tasks, regardless of whether video training was used.
Interestingly, even without video training, inference-time memory provides non-trivial gains across tasks without video annotations, suggesting the memory mechanism can compensate for the absence of temporal training signal when we distill the task-specialist $\F^k$ during unified training.

While video training drives temporal consistency, it is \textit{not} responsible for our per-task accuracy advantage over the unified image baselines.
Treating videos as independent images yields nearly identical per-task performance as shown in Appendix, confirming that the gains over baselines stem from the unified architecture and latent distillation rather than from video supervision.

\section{Conclusion}
\label{sec:conclusion}

Real-world data are inherently fragmented across incompatible domains.
In this work, we present \ours, learning unified video dense prediction from these disjoint sources.
It first learns task-specific embeddings on disjoint datasets, then distills them into a unified backbone via per-task latent projectors.
To bridge these domain gaps, we leverage the diffusion priors from internet-scale pretraining to enable generalization to unseen scene-task combinations.
\ours achieves competitive performance with per-task specialists and outperforms unified baselines across OOD scenarios with superior temporal and cross-task consistency.

\textbf{Limitations.}
Latent reconstruction is less effective for classification tasks, requiring additional fine-tuning to recover segmentation performance; future work could explore objectives that better preserve classification representations.

\textbf{Acknowledgements.}
This work was conducted in part during an internship at Adobe Research and supported in part by NSF (IIS-2144117).
Yihong Sun is supported by an NSF Graduate Research Fellowship.
We thank Yao-Chih Lee, Wei-Hong Li, Youming Deng, and Ashley P. Tsang for their valuable feedback.

\bibliographystyle{splncs04}
\bibliography{main}

\newpage
\appendix
\section{Implementation Details}

\subsubsection{VAE Encoder and Decoder.}
The VAE encoder $\Enc$ maps an input image $\x_i \in \RHW{3}$ into a compact latent code $\z_i = \Enc(\x_i) \in \Rhw{d}$, where the spatial downsampling factor is $8$ (\ie, $h = H/8$, $w = W/8$) and the latent channel dimension is $d=4$. Both $\Enc$ and $\Dec$ are frozen during all training stages.

\subsubsection{U-Net Backbone.}
No architectural changes are made to the pretrained U-Net.
As input to the U-Net, the clean image $\x_i$ is passed in along with a fixed text prompt (\texttt{"video\_dense\_pixel\_prediction"}).

\subsubsection{Pixel Projector $\PixProj^k$.}
The pixel projector $\PixProj^k$ maps the decoded feature map $\Dec'(\l_i^k) \in \RHW{d'}$ to the task-specific output $\hat{\y}_i^k \in \RHW{C^k}$, where $d'$ is the number of input channels to the final convolutional layer of the VAE decoder $\Dec$.
For all tasks, $\PixProj^k$ is a lightweight single $3{\times}3$ convolutional layer mapping $d'$ channels directly to $C^k$ outputs. 
All projector weights use Xavier uniform initialization and are trained from scratch during specialist training. 
Please refer to Table~\ref{tab:training_data} for $C^k$ for each task.

\subsubsection{Latent Projector $\LatProj^k$ (DPT Head).}
Each latent projector $\LatProj^k$ is a DPT-style head~\cite{ranftl2021vision} that fuses the final output feature with 3 intermediate features as follows from the unified backbone $\G$ and predict task-specific latent $\hat{\l}_i^k \in \Rhw{d}$.
\begin{itemize}[noitemsep]
    \item \texttt{mid\_block.attentions.0.transformer\_blocks.0}
    \item \texttt{up\_blocks.1.attentions.2.transformer\_blocks.0}
    \item \texttt{up\_blocks.2.attentions.2.transformer\_blocks.0}
\end{itemize}
The four feature maps are first projected to a common fusion dimension of $256$ via a $3{\times}3$ convolution and then progressively fused coarse-to-fine.
A final prediction head projects to the output latent dimension $d=4$.

\subsubsection{Extended Self-Attention Module.}
All self-attention blocks in the U-Net perform extended self-attention~\cite{streamv2v} when past frames are available in the memory bank.
The memory bank is never dequeued during training and enqueues/dequeues accordingly during inference. 
We further add a temporal positional embedding to differentiate the current frame from past frames.
As they are directly added to the past frame features, for tasks without video training data, the temporal positional embedding, which is zero-initialized, is not updated during training.
During inference, without learning, the positional embedding is effectively turned-off and reverts to the original extended self-attention~\cite{streamv2v}.

\section{Training Details}
\label{sec:training}

\subsection{Specialist Training}
\label{sec:training:specialist}

\subsubsection{Per-Task Loss Functions.}
Each specialist $(\F^k, \PixProj^k)$ is trained end-to-end with a task-specific loss $\Loss^k$ computed between the ground truths $\y_i^k$ and predictions $\hat{\y}_i^k = \PixProj^k(\Dec'(\F^k(\Enc(\x_i))))$ as follows:
\begin{enumerate}
    \item \textbf{Depth}: $\Loss^\text{Depth}$ contains (1) Scale-and-shift-invariant (SSI) loss applied to both images and videos with weight $1.0$; (2) a temporal gradient matching (TGM) loss~\cite{chen2025videodepthanything} with weight $1.0$; and (3) enforcing far-plane regularization on invalid pixels beyond the far-plane with weight $3.0$.
    $$\Loss^\text{Depth} = \Loss^\text{Depth}_\text{SSI} + \Loss^\text{Depth}_\text{TGM} + 3.0\,\Loss^\text{Depth}_\text{far}$$

    \item \textbf{Normal}: $\Loss^\text{Normal}$ contains (1) an angular loss~\cite{dsine} applied to images with weight $1.0$; and (2) a far-plane regularization on invalid pixels beyond the far plane with weight $3.0$.
    $$\Loss^\text{Normal} = \Loss^\text{Normal}_\text{ang} + 3.0\,\Loss^\text{Normal}_\text{far}$$

    \item \textbf{Albedo}: $\Loss^\text{Albedo}$ contains an MSE loss on the RGB output with weight $1.0$.
    $$\Loss^\text{Albedo} = \Loss^\text{Albedo}_\text{MSE}$$

    \item \textbf{Shading}: $\Loss^\text{Shading}$ contains an MSE loss on the RGB output with weight $1.0$.
    $$\Loss^\text{Shading} = \Loss^\text{Shading}_\text{MSE}$$

    \item \textbf{Boundary}: $\Loss^\text{Boundary}$ contains separate MSE losses on positive (boundary) and negative (non-boundary) pixels. The positive-pixel MSE has weight $1.0$ and the negative-pixel MSE has weight $20.0$, upweighting the sparse boundary signal.
    $$\Loss^\text{Boundary} = \Loss^\text{Boundary}_\text{pos-MSE} + 20.0\,\Loss^\text{Boundary}_\text{neg-MSE}$$

    \item \textbf{Semantic Segmentation}: $\Loss^\text{SemSeg}$ contains (1) a cross-entropy loss on per-pixel class logits with weight $1.0$; and (2) a TGM loss on video logit predictions with weight $3.0$.
    $$\Loss^\text{SemSeg} = \Loss^\text{SemSeg}_\text{CE} + 3.0\,\Loss^\text{SemSeg}_\text{TGM}$$

    \item \textbf{Material Segmentation}: $\Loss^\text{MatSeg}$ contains a cross-entropy loss with weight $1.0$.
    $$\Loss^\text{MatSeg} = \Loss^\text{MatSeg}_\text{CE}$$

    \item \textbf{Human Part Segmentation}: $\Loss^\text{PartSeg}$ contains (1) a cross-entropy loss with weight $1.0$ (ignore index $15$); and (2) a non-empty regularization term with weight $1.0$ to account for missing human part annotations in DensePose~\cite{densepose}.
    $$\Loss^\text{PartSeg} = \Loss^\text{PartSeg}_\text{CE} + \Loss^\text{PartSeg}_\text{reg}$$
\end{enumerate}

\subsubsection{Training Details.}
All models are trained with AdamW~\cite{loshchilov2017decoupled} with $\beta_1{=}0.9$, $\beta_2{=}0.999$, weight decay $10^{-2}$, $\epsilon{=}10^{-8}$, and gradient clipping at max norm 1.0.
Learning rate is $3{\times}10^{-5}$ with linear warm-up over 1000 steps, then linear decay over the full training duration.
All specialists, except for semantic and material segmentation, are trained for $30,000$ iterations with effective batch size $16$ and temporal clip length $T{=}4$.
Semantic segmentation and material segmentation specialists are trained for $80,000$ iterations due to the larger number of output classes and the cross-entropy classification objective.
Specialist training takes roughly $9.3$ hours per $10,000$ iterations on eight NVIDIA A100 GPUs.

\subsection{Unified Model Training}
\label{sec:training:unified}

The unified backbone $\G$ and latent projectors $\{\LatProj^k\}$ are trained with a union of all labeled data in Table~\ref{tab:training_data}, sampled with equal weight between image and video data.
In practice, we select the threshold for latent temporal gradient matching $\varepsilon$ as $2.0$, and TGM weight $\lambda$ as $5.0$ relative to the per-frame $\ell_1$ reconstruction loss.
Unified training takes roughly $15$ hours per $10,000$ iterations on $16$ NVIDIA A100 GPUs with effective batch size 32 and temporal clip length $T{=}4$.

\subsection{Segmentation Fine-Tuning}
\label{sec:training:finetune}

After unified training, we perform a lightweight fine-tuning step for semantic, material, and human part segmentation tasks that updates only the task-specific projectors $(\LatProj^k, \PixProj^k)$ while keeping $\G$ entirely frozen.
The fine-tuning uses the same task-specific datasets, loss functions, and optimizer settings as specialist training, with only $(\LatProj^k, \PixProj^k)$ trainable.
Fine-tuning runs for $30,000$ iterations, taking roughly $6.2$ hours per $10,000$ iterations on eight NVIDIA A100 GPUs.

\section{Evaluation Details}
\label{sec:eval}

\subsection{Evaluation Datasets and Metrics}
\label{sec:eval:datasets}

For Out-of-Distribution evaluations, we list our evaluation setups as below.
\begin{enumerate}
    \item \textbf{Depth.} We follow the evaluation protocol of Marigold~\cite{marigold}, applying least-squares scale-and-shift alignment and computing AbsRel and $\delta_1$.

    \item \textbf{Surface Normals.} We follow the evaluation protocol of DSINE~\cite{dsine}, $\ell_2$-normalizing predictions and computing mean angular error and the $11.25^\circ$ accuracy.

    \item \textbf{Albedo.} We follow the evaluation protocol of IIW~\cite{bell14intrinsic}, reporting Weighted Human Disagreement Rate (WHDR) at the $15\%$ thresholds. Lower WHDR is better.

    \item \textbf{Shading.} We follow the evaluation protocol of SAW~\cite{kovacs17shading}, reporting Precision at $70\%$ Recall (P@0.7) and Average Precision (AP).

    \item \textbf{Boundaries.} We derive instance boundary maps from the instance annotations of YT-VIS~\cite{ytvis} and Mapillary~\cite{mapillary}, and evaluate using the optimal-dataset-scale F-measure (odsF) following InvPT~\cite{invpt}.

    \item \textbf{Semantic Seg.} We evaluate on ADE20K~\cite{ade20k} using the existing COCO-to-ADE20K label mapping from MSeg~\cite{mseg}, reporting mIoU and fwIoU.

    \item \textbf{Material Seg.} We construct material labels for ADE20K~\cite{ade20k} by mapping semantic classes that are homogeneous in material (\eg \textit{lake} $\to$ \textit{water}, \textit{curtain} $\to$ \textit{fabric}), and evaluate mIoU and fwIoU within the mapped classes.

    \item \textbf{Human Part Seg.} We remap our 14 predicted body-part labels to the 6 coarse categories of Pascal-Human-Part~\cite{pascalhuman} (\ie \textit{head}, \textit{torso}, \textit{upper arms}, \textit{lower arms}, \textit{upper legs}, \textit{lower legs}), ignore background and ignore label, and report mIoU and fwIoU.
\end{enumerate}
In addition, semantic and material segmentation predictions are post-processed with a CRF solely for visualization; all quantitative evaluations are performed on raw model outputs.

\subsection{Temporal Consistency}
\label{sec:eval:temporal}

\subsubsection{Evaluation Metric: mVC-$t$.}
We extend Video Consistency (mVC-$t$)~\cite{miao2021vspw}, originally proposed for video semantic segmentation, to also measure temporal consistency for surface normals, albedo, shading, and instance boundaries.
In the original formulation, a pixel is ``stable'' across a $t$-frame window if all $t$ frames agree on the same class label; mVC-$t$ then measures the fraction of ground-truth-stable pixels that are also stable in the prediction.
We adapt this definition to non-semantic tasks as follows.
For instance boundaries, we first binarize the prediction at a fixed threshold before applying standard mVC-$t$.
For continuous-valued regression tasks, we define pairwise stability between consecutive frames using task-specific thresholds:
\begin{itemize}[noitemsep]
    \item \textbf{Surface Normals:} A pixel is stable if the angular difference between consecutive frames is ${<}11.25^\circ / 2$.
    \item \textbf{Albedo and Shading:} A pixel is stable if the mean $\ell_1$ difference across channels between consecutive frames is ${<}0.1$, with values normalized to $[0, 1]$.
\end{itemize}

\subsubsection{Evaluation Datasets.}
Depth is evaluated on ScanNet~\cite{scannet} video evaluation metrics (AbsRel, $\delta_1$).
Surface normals, albedo, and shading are evaluated on InteriorNet~\cite{interiornet} video sequences, where we randomly sample $120$ videos each containing $100$ frames.
Instance boundaries are evaluated on YT-VIS~\cite{ytvis}.
Semantic segmentation is evaluated on VIPSeg~\cite{vipseg}.

\subsection{Cross-Task Consistency Metrics}
\label{sec:eval:cross}

\subsubsection{Depth $\leftrightarrow$ Normal.}
To measure consistency between depth and normal, we compute the error between surface normals derived from depth predictions and the directly-predicted surface normals. 
Specifically, we compute the gradient of the Gaussian-smoothed ($\sigma{=}3.0$) depth surface with respect to pixel coordinates using camera intrinsics.
Then, we compute mean angular error and $11.25^\circ$ accuracy between the depth-derived normal and the predicted surface normal on ScanNet~\cite{scannet}.

\subsubsection{Human Part $\leftrightarrow$ Semantic.}
For consistency between human part and semantic segmentation, we measure the \% inclusion of predicted human-part pixels within the predicted human semantic category.
Let $\mathcal{H}{=}\{p : \text{partseg}(p){>}0\}$ denote the set of pixels predicted as any human part label, and $\mathcal{S}{=}\{p : \text{semseg}(p){=}c_\text{person}\}$ the set of pixels predicted as the ``person'' class.
We report $\%\,\text{inclusion} = |\mathcal{H} \cap \mathcal{S}| / |\mathcal{H}|$, evaluated on COCO validation images with human-part annotations. 

\subsubsection{Albedo $\leftrightarrow$ Shading.}
To measure intrinsic consistency, we directly measure reconstruction performance of the diffuse image from predicted albedo and shading. Specifically, we compute how well the product $\hat{\x} = \hat{\mathbf{a}} \odot \hat{\mathbf{s}}$ (predicted albedo $\times$ predicted shading, clipped to $[0,1]$) reconstructs the ground-truth diffuse image, using PSNR and LPIPS on Hypersim~\cite{hypersim}.

\subsection{Additional Results}

\subsubsection{Per-task performance without video training.}
To isolate the source of \ours's per-task accuracy advantage over the unified image baselines, we disable video training and inference entirely, treating all videos as independent images (denoted as UniD-Image).
As shown in Table~\ref{tab:supp:image_abl}, UniD-Image attains comparable per-task performance to the full \ours across all eight tasks, with differences within evaluation noise on most tasks.
This confirms that the gains over baselines stem from the unified architecture and latent distillation, whereas video training instead drives temporal consistency, as analyzed in the main paper.

\begin{table}[t]
\vspace{-0.25cm}
\centering
\caption{Multi-Task Performance. Best is bolded.}
\resizebox{\textwidth}{!}{%
\begin{tabular}{l@{\hspace{1pt}}  c@{\hspace{1pt}}c@{\hspace{1pt}}c@{\hspace{1pt}}c@{\hspace{1pt}}c@{\hspace{1pt}}c@{\hspace{1pt}}c@{\hspace{1pt}}c}
\toprule

  & \textbf{Depth}
  & \textbf{Normal}
  & \textbf{Albedo}
  & \textbf{Shading}
  & \textbf{Boundary}
  & \textbf{Semantic}
  & \textbf{Material}
  & \textbf{Part} \\
\cmidrule(lr){2-9}
 Method & NYUv2 & NYUv2 
 & IIW & SAW 
 & Map  & CS 
 & ADE  & Pascal \\
  & $\text{AbsRel}\downarrow$ & $\text{mean}\downarrow$
 & $\text{WHDR}\downarrow$ & $\text{AP\%}\uparrow$
 & $\text{odsF}\uparrow$ & $\text{mIoU}\uparrow$
 & $\text{mIoU}\uparrow$ & $\text{mIoU}\uparrow$ \\
\midrule
UniD-Image & 0.060 & \textbf{16.0} & 0.210 & 92.5 & \textbf{61.6} & \textbf{58.9} & \textbf{67.1} & \textbf{66.0}   \\
UniD & \textbf{0.059} & 16.2 & \textbf{0.207} & \textbf{92.6} & 57.2  & 58.4 & 66.0  & 65.4 \\
\bottomrule
\end{tabular}
}
\label{tab:supp:image_abl}
\end{table}

\begin{table}[t]
\centering
\caption{In-Domain Evaluation of Surface Normal and Instance Boundary Estimation on InteriorNet~\cite{interiornet} and YT-VIS~\cite{ytvis}.}
\resizebox{0.6\textwidth}{!}{%
\begin{tabular}{@{\hspace{4pt}}c@{\hspace{4pt}}l c@{\hspace{4pt}}c@{\hspace{4pt}}c}
\toprule
\multirow{3}{*}{} & \multirow{3}{*}{Method} & \multicolumn{2}{c}{\textbf{Normal}} & \textbf{Boundary} \\
 & & \multicolumn{2}{c}{InteriorNet} & YT-VIS \\
\cline{3-5}
 & & $\text{mean} \downarrow$ & $11.25^\circ$$\uparrow$ & $\text{odsF} \uparrow$ \\
\midrule
\multirow{2}{*}{\rotatebox[origin=c]{90}{Spec.}}
& NormalCrafter~\cite{bin2025normalcrafter} & 12.1 & 76.4 & - \\
& \ours ($\F^k$) & 14.9	& 63.8 & 78.5 \\
\midrule
\multirow{3}{*}{\rotatebox[origin=c]{90}{Unified}}
& DICEPTION~\cite{diception}       & 17.3 & 70.3 & - \\
& DINOv3-H~\cite{dinov3}        & 11.7	& 76.5 & 79.4 \\
& \ours ($\G$)    & 15.1	& 63.7 & 78.3 \\
\bottomrule
\end{tabular}
}
\label{tab:supp:perf}
\end{table}

\subsubsection{Performance Results on InteriorNet and YT-VIS}
For completeness, please refer to Table~\ref{tab:supp:perf} for performance results on InteriorNet~\cite{interiornet} and YT-VIS~\cite{ytvis}, where temporal consistency is reported in the main paper (Table~\ref{tab:temp_const}).
In addition, qualitative results on InteriorNet~\cite{interiornet} can be found in the visualization at \url{\projectpage}, under ``In-Domain Data".

With similar performances on in-domain benchmarks, \ours outperforms DINOv3-H~\cite{dinov3} substantially in OOD conditions as shown in Table~\ref{tab:temp_const} in the main paper.

\end{document}